\pdfoutput=1
\documentclass[runningheads]{llncs}
 
\usepackage{eccv}

\usepackage{eccvabbrv}

\usepackage{graphicx}
\usepackage{booktabs}
\usepackage{xcolor}
\usepackage{multirow}    
\usepackage{subcaption}
\definecolor{mygreen}{RGB}{0,140,0}
\definecolor{myred}{RGB}{180,0,0}
\usepackage{enumitem}
\newcommand{\specialcell}[2][c]{%
\begin{tabular}[#1]{@{}c@{}}#2\end{tabular}}

\usepackage[accsupp]{axessibility}  

\definecolor{lightorange}{RGB}{247, 228, 216}
\definecolor{lightblue}{RGB}{176, 226, 230}
\definecolor{lightblue}{RGB}{227 246 255}
\definecolor{lightgray}{RGB}{211 211 211}
\definecolor{purple}{RGB}{94,60,153}
\definecolor{orange}{RGB}{250,190,155}
\definecolor{darkblue}{HTML}{4a95f0}
\definecolor{darkpurple}{HTML}{c07be3}
\definecolor{darkgreen}{HTML}{30c246}
\definecolor{mygreen}{RGB}{0,140,0}
\definecolor{myred}{RGB}{180,0,0}
\definecolor{myblue}{HTML}{FDF5E0}

\PassOptionsToPackage{table,xcdraw}{xcolor}
\usepackage{amsmath}
\usepackage{algorithm}
\usepackage{algpseudocode}
\usepackage{amssymb}
\usepackage{diagbox}
\usepackage{booktabs}
\usepackage{colortbl}
\usepackage{graphicx}
\usepackage{wrapfig}
\usepackage{soul}
\usepackage[numbers]{natbib}
\usepackage[most]{tcolorbox} 
\usepackage{mdframed}
\usepackage{tabularx}
\usepackage{makecell}
\usepackage{array}
\usepackage{soul}
\usepackage[normalem]{ulem}
\usepackage{titletoc}

\newcommand{\comploss}{\emph{composition loss }}

\usepackage{hyperref}

\usepackage{orcidlink}

\begin{document}

\title{The {\textcolor{darkpurple}{A}\textcolor{darkblue}{R}\textcolor{darkgreen}{T}} of Composition: {\textcolor{darkpurple}{A}ttention-\textcolor{darkblue}{R}egularized \textcolor{darkgreen}{T}raining for Compositional Visual Grounding}}


\titlerunning{CompART}
\author{Jiayun Luo\inst{1} \and
 Mir Rayat Imtiaz Hossain\inst{1} \and
Pritam Sarkar\inst{1} \and Boyang Li\inst{2} \and
Leonid Sigal\inst{1}}
\authorrunning{Jiayun L et al.}

\institute{The University of British Columbia, Vancouver, BC V6T 1Z4, Canada \\
\email{\{letitial, rayat137, psarka03, lsigal\} @cs.ubc.ca}\\
\and
Nanyang Technological University, Singapore 639798\\
\email{boyang.li@ntu.edu.sg}}




\maketitle

\begin{abstract}
Vision–Language Models (VLMs) have achieved strong performance on (implicit and explicit) visual grounding and related tasks. However, such abilities are generally tested on simple, single-object phrases. We find that grounding performance degrades for complex, multi-object references. These limitations largely arise from training objectives that leverage image-caption alignment, where direct multi-object references are rare, number of possible such references is theoretically large (exponential in the number of objects) and attribution is difficult. To address this, without requiring any additional annotations, we propose Compositional Attention-Regularized Training (CompART) training, which decomposes captions into object-centric phrases and constructs composite phrases by pairing them with conjunctions. We then introduce a \comploss  that encourages the attention induced by a composite phrase to equal the sum of the attentions of its constituent phrases, promoting balanced multi-object localization. We evaluate CompART across four VLM architectures, spanning both contrastive-based and generative-based models, on four benchmarks for multi-object grounding and two VQA benchmarks for general visual understanding. 
CompART consistently achieves improved grounding for both single- and multi-object references across diverse VLM architectures and datasets, and further demonstrates enhanced visual understanding, as evidenced by gains on VQA, despite not being explicitly trained for this task. \footnote{We will release all models, data and code upon acceptance}

\end{abstract}

\section{Introduction}
\label{sec:intro}


\begin{figure}[t]
    \centering
    
    \includegraphics[width=1\linewidth,height=0.45\textheight,keepaspectratio]{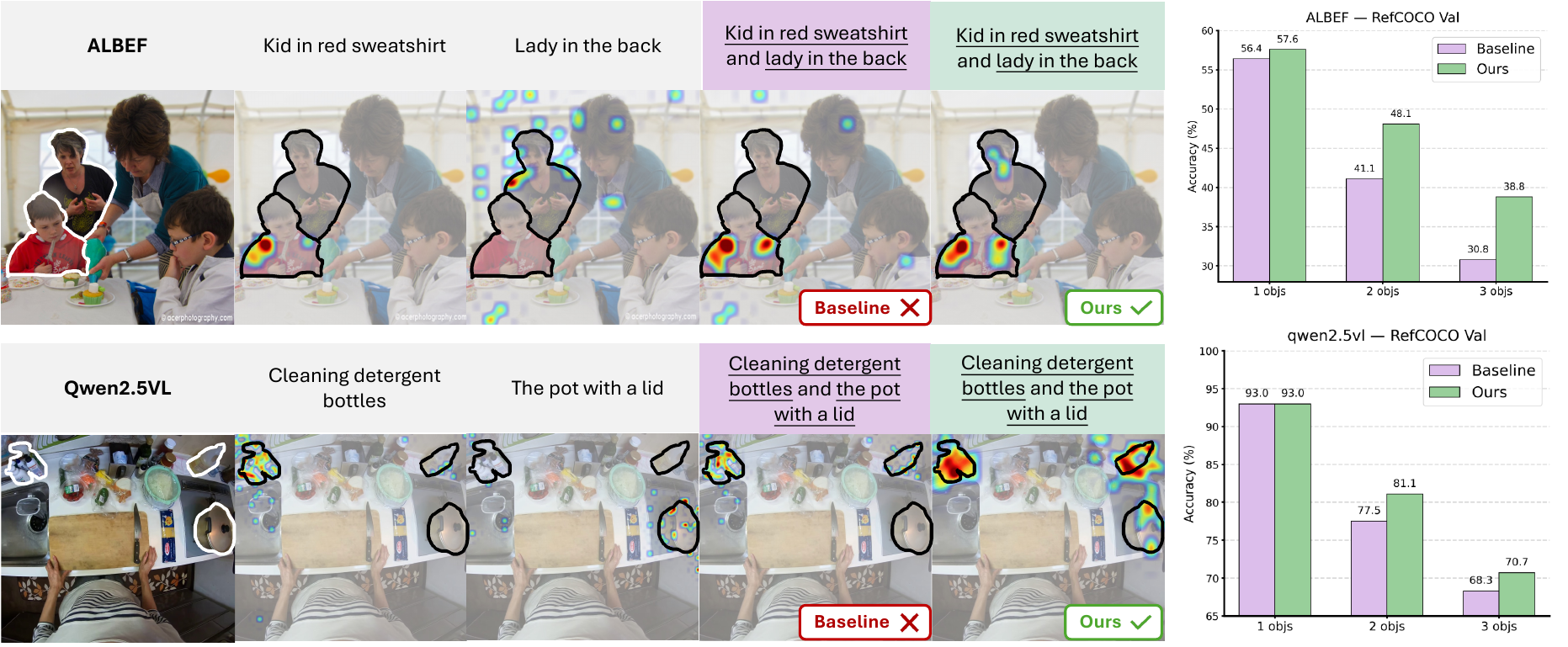}
    \vspace{-0.2in}
    \caption{
    We present qualitative and quantitative results of ALBEF and Qwen2.5VL on single- and multi-object grounding. While current VLMs perform well on simple single-object grounding tasks, their performance degrades substantially when localizing the same set of objects jointly using composite phrases (\colorbox{purple!30}{purple}). Proposed 
    CompART significantly improves performance (\colorbox{darkgreen!30}{green}), without requiring added annotations. 
    }
    
    \label{fig:motivation}
    \vspace{-0.15in}
\end{figure}


Vision–Language Models (VLMs) \cite{FLAVA,OSCAR-2020,BLIP,li2023blip2,bridgetower,flamingo, ViLT, mplug, LLAVA, clip-2021, ALIGN-2021, zhou2024tinyllava, bai2025qwen2} have demonstrated remarkable success in a wide range of vision tasks including, visual question answering (VQA)~\cite{VQA}, as well as both implicit/emergent \cite{PNPOVSS, SelfEQ, proxyclip, Composite}, where the grounding capability emerges from coarse image-caption supervision, and explicit visual grounding~\cite{chen2023shikra, you2023ferret, groundingdino, rasheed2024glamm}, where the models are explicitly trained to do grounding.
However, current VLMs continue to struggle with grounding complex queries involving multiple objects. As shown in Figure~\ref{fig:motivation}, when VLMs are prompted with a multi-object query such as {\tt `kid in red sweatshirt and lady in the back'}, attention often collapses onto a single salient region, {\tt `kid'}. As a result, grounding performance drastically degrades as the number of objects in the query increases (see Figure~\ref{fig:motivation}, \colorbox{purple!30}{purple bar}). This gap likely stems from the current training paradigm, which relies on unstructured image–caption pairs for vision-language alignment. In such data, direct multi-object references are relatively rare, the number of possible combinations grows exponentially with the number of objects, and attribution across objects is inherently ambiguous. Consequently, this limits VLMs' ability to attend to or localize multi-object phrases which can adversely affect model performance in tasks that require multi-object reasoning, such as 
grounding~\cite{liu2023gres, groundingdino, hu2023beyondonetone}, instance counting~\cite{seedbench}, and spatial reasoning~\cite{kamath2023s, chen2024spatialvlm}.

One solution for this limitation would be to collect dedicated data with composite references and corresponding images, however, this would require annotation of a potentially exponential\footnote{Exponential in the number of entities involved.} set of phrases and also would rely on VLMs implicitly learning compositional behavior. Instead, we explore whether the compositional structure inherent in captions can provide useful supervisory signals.
Prior work in natural language processing has shown that leveraging syntactic and compositional structure improves semantic modeling across tasks such as translation~\cite{andreas2013semantic, eriguchi2017learning} and sentence classification~\cite{komninos2016dependency, wu2020phrase2vec}, and earlier vision–language systems similarly exploited structured language representations for retrieval~\cite{vendrov2015order} and single object grounding~\cite{xiao2017weakly}. Related work in vision has also explored how structuring interactions between regions can improve the compositionality of visual representations~\cite{hossain2024framework, chen2019graph}. However, such structured supervision has yet to be integrated into modern large-scale VLM training pipelines. Motivated by this, we propose Compositional Attention Regularization training (CompART), a simple method that incorporates attention regularization, 
based on compositional text structure, into VLM optimization.
CompART improves model performance on multi-object grounding (Figure \ref{fig:motivation} \colorbox{darkgreen!30}{green bar}), instance counting and spatial relation understanding, without requiring added human annotations. 

Specifically, we first decompose each image caption into \emph{object-centric phrase-chunks}, each corresponding to a distinct visual entity, and then construct \emph{composite phrases} by pairing two such phrases using a simple conjunction (``and''). 
Motivated by the observation that VLMs reliably localize objects when queried individually but struggle with multi-object queries, we introduce \comploss  that exploits the emergent localization behavior of VLMs on single-object queries~\cite{PNPOVSS, SelfEQ, kang2025your}. Concretely, our \comploss encourages the attention map induced by a composite phrase to approximate the sum of the attention maps of its constituent phrases, thereby explicitly regularizing attention. As a result, \comploss mitigates single-object dominance and promotes simultaneous multi-object grounding. Although our \comploss  is applied only to object pairs during training due to computational constraints, the resulting model generalizes to grounding scenarios involving more than two-objects (see Figure \ref{fig:motivation}) and further demonstrates improvements across diverse VQA tasks.




We evaluate CompART across four representative VLM architectures, spanning both contrastive models (ALBEF~\cite{ALBEF} and BLIP~\cite{BLIP}) and generative models (TinyLLaVA~\cite{zhou2024tinyllava} and Qwen2.5VL~\cite{bai2025qwen2}). We include contrastive VLMs alongside more recent generative models because, despite having significantly fewer parameters, they often exhibit comparable grounding performance~\cite{kang2025your,bhatia2024local,chen2023shikra}, making them particularly attractive in low-compute settings.
We assess the impact of CompART on six benchmarks, comprising four grounding benchmarks and two VQA benchmarks. Across all settings, CompART consistently improves over the corresponding base models. For example, on average, CompART improves two-object grounding performance of TinyLLaVA-0.5B by 4.9\% points and Qwen2.5VL by 2.6\% points across all four grounding benchmarks. 
Importantly, our method introduces no architectural modifications to the base models. Instead, we propose a compositional attention regularization objective that is architecture-agnostic and integrates seamlessly into standard training pipelines, requiring no additional dense or bounding-box annotations. Despite its simplicity, CompART enhances multi-object grounding and improves broader visual understanding, as reflected by consistent gains across benchmarks.

\vspace{0.1in}
\noindent
{\bf Contributions.} In summary, we make the following contributions:

\vspace{0.1in}
\noindent
\textbf{(1)} identify a consistent performance gap when transitioning from single- to multi-object grounding in both contrastive and generative VLMs;

\vspace{0.05in}
\noindent
\textbf{(2)} introduce Compositional Attention Regularization Training (CompART), a simple 
dense-annotation-free solution for improving multi-object grounding;

\vspace{0.05in}
\noindent
\textbf{(3)} demonstrate that proposed framework leads to consistent improvements across diverse VLMs on multi-object grounding and VQA benchmarks. 

\section{Related Work}
\label{sec:relate}


\noindent\textbf{Vision–Language Models.} Vision–language pretraining underpins modern multimodal learning and achieves strong performance across diverse tasks. Existing VLMs broadly fall into two categories: (1) {\em contrastive-based models}, which learn aligned image–text representations via large-scale contrastive objectives, including dual-encoder approaches such as CLIP~\cite{clip-2021} and ALIGN~\cite{ALIGN-2021}, as well as fusion-based variants such as ALBEF~\cite{ALBEF} and BLIP~\cite{BLIP} that enable explicit cross-modal interaction; and (2) {\em generative-based} VLMs that integrate visual encoders with large language models (e.g., LLaVA~\cite{LLAVA}, Phi3.5V~\cite{abdin2024phi3technicalreporthighly}, DeepseekVL~\cite{lu2024deepseek}, Molmo~\cite{deitke2025molmo}, InternVL~\cite{zhu2025internvl3}, TinyLLaVA~\cite{zhou2024tinyllava}, Qwen2.5-VL~\cite{bai2025qwen2}), enabling open-ended multimodal reasoning and generation.

\vspace{0.05in}
\noindent \textbf{Visual Grounding.} Visual grounding methods can broadly be categorized into implicit and explicit grounding approaches. Implicit grounding~\cite{SelfEQ, jin2023refclip, Composite, hossain2026segmentation} refers to methods where phrase–region correspondences emerge from a model’s internal representations, such as cross-modal attention maps, without directly predicting spatial outputs. In contrast, explicit grounding~\cite{rohrbach2016grounding, yu2018mattnet,lu2019vilbert, lai2024lisa,rasheed2024glamm} approaches train models to localize objects by predicting spatial outputs such as bounding boxes, coordinates, or segmentation masks directly.

Several implicit grounding approaches are trained in a weakly supervised manner using only paired image–text data without relying on region proposals or detector supervision. These methods encourage phrase–region alignment through training strategies applied to vision–language models, such as progressive learning~\cite{progresssive_comprehension}, consistency across paraphrases~\cite{lee2023weakly, SelfEQ}, and object disentanglement constraints~\cite{Composite}.
Another line of work exploits pseudo ground-truth signals such as region proposals generated by object detectors, where textual phrases are aligned with candidate regions extracted from the image~\cite{rohrbach2016grounding, yu2018mattnet, lu2019vilbert, chou2022semi, luo2024apl, jin2023refclip}. These methods learn phrase–region correspondences by matching referring expressions to candidate region proposals. Further improvements have been explored through strategies such as knowledge distillation from teacher models~\cite{wang2021improving}, prompt-based optimization~\cite{luo2024apl}, and curriculum learning~\cite{dai2024curriculum}.
While effective, these approaches primarily enhance contrastive-based models for single-object grounding and often rely on task-specific architectural modifications. 





More recent work, such as GroudingDINO~\cite{groundingdino}, GRES~\cite{liu2023gres} and Beyond-one-to-one\cite{hu2023beyondonetone}, leverages dense annotations to extend contrastive models to more challenging multi-object grounding scenarios. ReMeREC~\cite{hu2025remerec} further introduces relation-aware multi-entity referring comprehension to explicitly model inter-object relationships. In parallel, several grounding-oriented multimodal large language models extend instruction-tuned LLMs with explicit spatial grounding capabilities, enabling direct coordinate prediction or mask generation via segmentation decoders ~\cite{lai2024lisa,rasheed2024glamm,zhang2024llavagrounding,zhang2024groundhog,chen2023shikra,you2023ferret,yin2025rodmllm,jiang2025rexthinker}. However, these methods largely depend on dense annotations to support multi-object grounding.

Different from the above approaches, CompART leverages caption compositionality and attention regularization to improve multi-object grounding in both contrastive-based and generative-based VLMs. Our method is simple, architecture-agnostic, and does not require dense annotations.

\noindent \textbf{The Role of Grounding in Visual Understanding.} Accurate visual grounding has been shown to improve broader visual understanding tasks such as VQA \cite{reich2024role}. Several studies demonstrate that encouraging models to attend to question-relevant image regions leads to improved reasoning accuracy and robustness~\cite{selvaraju2019taking, wu2019self, le2023guiding}. Other works explicitly learn grounded representations or analyze the role of grounding in VQA, showing that aligning language queries with relevant visual regions improves interpretability and model performance~\cite{urooj2021found, reich2024role}.

\begin{figure}[!t]
\centering
  \includegraphics[width=0.95\linewidth]
{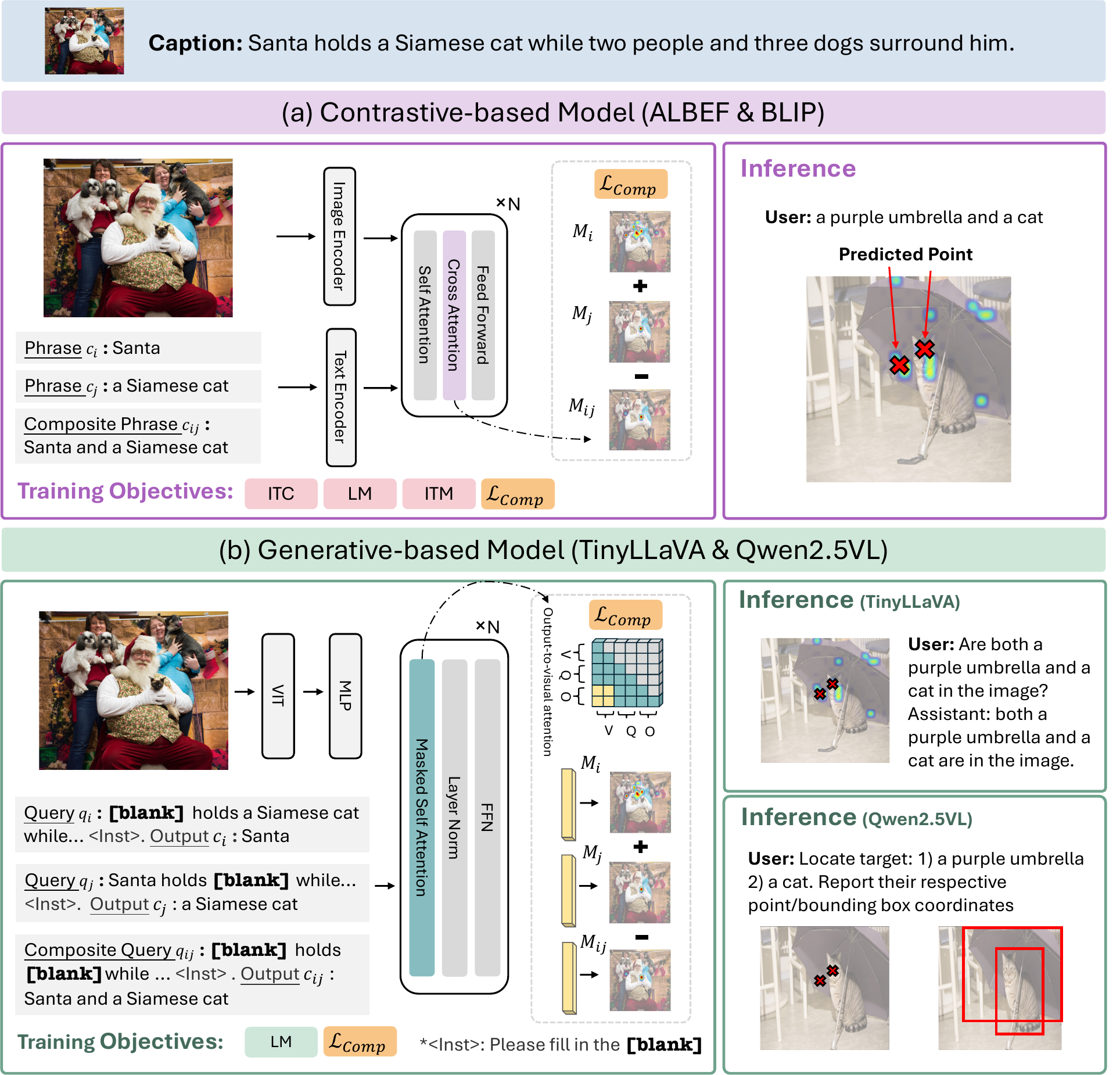}
    \caption{\textbf{Overview of CompART.}     
    During training, we first extract \emph{object-centric phrase-chunks} from the image caption and form \emph{composite phrases} by pairing them. Then, we extract localization maps for the phrases. Since contrastive and generative VLMs are trained differently, the composite queries are formulated slightly differently for each VLM. Our proposed \comploss ($\mathcal{L_\text{Comp}}$) encourages the localization map of a composite phrase to correspond to the union of the object regions associated with the phrase chunks. 
    During inference, for models without explicit grounding capability (ALBEF, BLIP, and TinyLLaVA), we use the peak of the attention maps corresponding to the query objects as the predicted points. 
    Since Qwen2.5VL supports explicit grounding capabilities, we directly obtain output points or bounding box coordinates.} 
    \label{fig:main}
    \vspace{-10pt}
\end{figure}

\section{Compositional Attention Regularization Training (CompART)}
\label{sec:compart}



In Figure~\ref{fig:main}, we present a high-level overview of CompART. We first extract \emph{object-centric phrase-chunks}  from the original image captions and form \emph{composite phrases} by pairing them (Sec.~\ref{subsec:phrasechunking}). We then compute localization maps for both the individual phrases and their composite counterparts (Sec.~\ref{subsec:localization}), which are used to derive the \emph{composite loss} for attention-regularized training (Sec.~\ref{subsec:comploss}). The inference pipeline is discussed in Sec.~\ref{subsec:inference}. We apply CompART to both contrastive-based ({\em i.e.}, ALBEF~\cite{ALBEF}, BLIP~\cite{BLIP}) and generative-based models ({\em i.e.}, TinyLLaVA~\cite{zhou2024tinyllava}, Qwen2.5VL~\cite{bai2025qwen2}). 
In addition to TinyLLaVA and Qwen2.5VL, which inherently compute text-to-visual attention by virtue of their decoder-only LLM architecture, we also consider contrastive-based models such as ALBEF and BLIP, as they provide explicit text-conditioned visual attention that enables phrase-level localization.
However, due to their architectural differences, we adopt slightly different strategies as described below.




\subsection{Phrase Chunking}
\label{subsec:phrasechunking}

\noindent \textbf{Contrastive-based Models.}
Given an image $v$ and its corresponding caption $c$, we employ an LLM to extract object-centric phrase chunks $c_i$ and $c_j$, where $i$ and $j$ refer to the $i$-th and $j$-th phrases in $c$. We then construct a composite phrase chunk $c_{ij}$ by combining $c_i$ and $c_j$.
Subsequently, the phrase chunks are paired with $v$, which serves as the model input. See examples in Figure~\ref{fig:main}(a).

\vspace{0.05in}
\noindent \textbf{Generative-based Models.} 
Similar to the aforementioned procedure, 
for an image $v$ and its corresponding caption $c$, we extract object-centric phrase chunks $c_i$ and $c_j$. Since generative-based VLMs produce outputs autoregressively conditioned on the input image and query, we construct a query $q_i$ by replacing $c_i$ in $c$ with a special token {\small \tt [blank]} and appending the instruction ``Please fill in the {\small \tt [blank]}.'' The removed phrase $c_i$ serves as the ground-truth target. Similarly, we obtain $q_j$ by replacing $c_j$ in $c$ with {\small \tt [blank]}. To form the query $q_{ij}$ corresponding to the composite phrase $c_{ij}$, we replace both $c_i$ and $c_j$ in $c$ with {\small \tt [blank]} and use $c_{ij}$ as the target output. See examples in Figure~\ref{fig:main}(b).


\subsection{Phrase Localization} 
\label{subsec:localization}



\noindent \textbf{Contrastive-based Models:} 
For a given image–phrase pair $\{v, c_i\}$, we obtain the text-to-image cross-attention map $\mathbf{A}_i \in \mathbb{R}^{K \times P}$, where $P$ denotes the total number of visual patches and $K$ denotes the number of tokens corresponding to $c_i$.  However, $\mathbf{A}_i$ is often noisy and leads to over-segmentation~\cite{PNPOVSS}. To address that, following~\cite{PNPOVSS, SelfEQ}, we adopt a GradCAM-based refinement strategy~\cite{GradCAM2017} which suppresses spurious activations and emphasizes regions most relevant to the phrase. 
Specifically, the refined localization map is obtained as the element-wise product between $\mathbf{A}_i$ and the gradient of the image--text matching (ITM) loss $\mathcal{L}_{\text{ITM}, i}$ with respect to $\mathbf{A}_i$\footnote{
While ALBEF and BLIP are also trained with image-text-contrastive (ITC) and language-modelling (LM) objectives, we compute gradients specifically with respect to ITM, as it directly enforces matching between the image and phrase \textit{within} a sample. In contrast, ITC is a contrastive objective applied \textit{across} samples, and LM operates purely on language.}.
Subsequently, we obtain the phrase-specific localization map $\mathbf{M}_i \in \mathbb{R}^{P}$ by averaging over $K$ tokens as:
\begin{equation}
\label{equation:Gradcam}
\mathbf{M}_i =
\frac{1}{K}\sum_{k=1}^{K}
\texttt{ReLU}\!\left(
\mathbf{A}_i^{(k)} \cdot
\frac{\partial \mathcal{L}_{\text{ITM}, i}}{\partial \mathbf{A}_i^{(k)}}
\right).
\end{equation}

\noindent \textbf{Generative-based Models:} 
For a given image–phrase pair $\{v, c_i\}$, 
we construct the model input as $\{v, q_i\}$ and set the target output to $c_i$, as defined in Sec.~\ref{subsec:phrasechunking}.
For phrase localization, we extract the output-to-visual ($c_i$ to $v$) attention as $\mathbf{A}_i\in R^{K \times P}$, where $K$ represents the number of tokens in $c_i$ and $P$ represent the total number of patches in $v$. We define our phrase-specific localization map $\mathbf{M}_i \in \mathbb{R}^{P}$ as:
\begin{equation}
\label{equation:Gradcam1}
\mathbf{M}_i =
\frac{1}{K}\sum_{k=1}^{K}\!\left(
\mathbf{A}_i^{(k)}
\right).
\end{equation}

\subsection{Composition Loss}
\label{subsec:comploss}

We next introduce \comploss  ($\mathcal{L}_\text{Comp}$) for attention-regularized training to improve visual grounding.
$\mathcal{L}_\text{Comp}$ encourages the model to jointly attend to multiple object regions associated with each individual phrase chunks by minimizing the Euclidean distance between the composite phrase attention map ($\mathbf{M}_{ij}$) and the sum of the individual phrase attention maps ($\mathbf{M}_i, \mathbf{M}_j$), defined as:
%
\begin{equation}
\mathcal{L}_{\text{Comp}}
=
\left\|
\mathbf{M}_i
+
\mathbf{M}_j
-
\mathbf{M}_{ij}
\right\|_2^2 .
\end{equation}


\noindent
$\mathcal{L}_\text{Comp}$ directly addresses the lack of multi-object grounding by explicitly regularizing consistency between the attention maps of individual phrase chunks and their composite form.





\subsection{Final Objective} 


Finally, we train the models using a weighted sum of their respective base training objectives ($\mathcal{L}_{\text{base}}$) and our \comploss  ($\mathcal{L}_\text{Comp}$). 
ALBEF and BLIP are trained with their base objectives, such as
ITC, ITM, and LM losses, whereas TinyLLaVA and Qwen2.5VL only use LM loss. The final training objective is defined as:
\begin{equation}
\mathcal{L}_{\text{total}} = \mathcal{L}_{\text{base}} + \lambda \times \mathcal{L}_{\text{Comp}}, 
\end{equation}
where $\lambda$ is a loss coefficient 
to scale the $\mathcal{L}_{\text{Comp}}$ to be the same range as $\mathcal{L}_{\text{base}}$. 
The combination of base and \comploss  encourages the models to retain their general vision-language capabilities while improving visual grounding.



\subsection{Inference}\label{subsec:inference}






The goal of CompART is to enhance the overall visual grounding capabilities of VLMs, rather than explicitly optimizing them for specific VQA tasks. Nevertheless, stronger visual grounding is expected to translate into improved performance on general VQA benchmarks. In the following, we describe our inference strategies for both visual grounding and VQA tasks.

\noindent \textbf{Visual Grounding:}
Ability and protocol for obtaining visual grounding differs among VLM families. 
For example, 
Qwen2.5-VL~\cite{bai2025qwen2}, which is instruction-tuned for coordinate prediction, has an {\em explicit} ability to localize objects, and one can directly instruct the model to output object points or bounding boxes coordinates. 
ALBEF~\cite{ALBEF}, BLIP~\cite{BLIP} and TinyLLaVA~\cite{zhou2024tinyllava}, are not equipped with such abilities and only posses {\em implicit} grounding typically exhibited by their attention maps \cite{SelfEQ, PNPOVSS, proxyclip, Composite}. 
Prior works either (i) post process such maps to produce segmentations \cite{PNPOVSS,hossain2026segmentation,cha2023learning} or (ii) evaluate such models using a pointing metric \cite{SelfEQ, progresssive_comprehension, TRIS, wwbl, g++, TAS}, which measures ability of the model to produce a coordinate (typically peak of the attention map) within the referred object's bounding box. 
However, the first approach is sensitive to the choice of post-processing methods \cite{densecrf,pamr}, potentially obfuscating grounding abilities. 
We adopt the second mechanism for such models and derive localization map from the text-to-image attention maps following prior work \cite{SelfEQ, kang2025your} and as described in Sec. \ref{subsec:localization} --  treating the peak of the map as the prediction of the object location.



\noindent
\textbf{VQA:} We follow the default inference setup of the base models for VQA tasks. 

\section{Experiments \& Results}
\label{exp}

\subsection{Training Details}
\label{subsec:impdetail}

We use ALBEF, BLIP, TinyLLaVA-0.5B, and Qwen2.5VL-3B as base VLMs to implement CompART. Our training data is sourced from two popular image-captioning datasets: Pixmo Caption~\cite{deitke2025molmo} and MSCOCO~\cite{COCO}. Specifically, we obtain 8.5K random image–caption pairs from Pixmo Caption, followed by using Qwen3-30B~\cite{yang2025qwen3} to prepare the object-centric phrase chunks. We construct an average of around 10 composite phrases per image, resulting in a total of 90K training samples. In the case of MSCOCO, we directly use object-centric phrase chunks from \cite{SelfEQ}, which contains 123K images and an average of 11 composite phrases per image, resulting in a total of 1.4M training samples. Both Pixmo and MSCOCO are used to train ALBEF, BLIP, and TinyLLaVA; however, due to computational constraints, we only use Pixmo to train Qwen2.5VL. 

We calculate object localization maps from the grounding layers of the VLMs. Specifically, for ALBEF and BLIP, we follow \cite{SelfEQ,PNPOVSS} and extract text-to-image attention maps from the 3$^{\text{rd}}$ and 9$^{\text{th}}$ cross-attention layers, respectively. For TinyLLaVA-0.5B and Qwen2.5VL-3B, we extract output-to-visual attention maps from the 23$^{\text{rd}}$ layer and the 21$^{\text{st}}$ layers of the VLMs. These layers are chosen empirically, as they assign the highest attention to visual tokens. All experiments are conducted on 8 NVIDIA L40 GPUs (48GB each). Full training details and hyperparameters are provided in the supplementary.

\subsection{Evaluation Setup}

\noindent \textbf{Visual Grounding.} We evaluate CompART on four visual grounding benchmarks: Flickr30K~\cite{flickr30k}, along with all validation and test splits of RefCOCO~\cite{refcoco+}, RefCOCO+~\cite{refcoco+}, and RefCOCOg~\cite{refcocog}. 
We assess performance across single- to five-object grounding settings. For single-object grounding, we use the full evaluation splits. 
For multi-object grounding, we consider images that contain ground-truth annotations for multiple objects and evaluate all possible object combinations. Specifically, for a $k-$object grounding setting, we use images that contain at least $k$ annotated objects and evaluate all possible $k-$object combinations within each image. Detailed evaluation set statistics are provided in the supplementary material; data splits will be released upon publication.



It should be noted that for multi-object grounding, predicting a single point is insufficient: a prediction is considered correct if it falls within any one of the target objects, which does not reflect whether the model attends to all referred objects. Similarly, when predicting a single bounding box and evaluating it using Acc@0.5, the metric does not capture the extent to which the model simultaneously localizes multiple objects. Therefore, multi-object grounding evaluation requires predicting distinct points or bounding boxes for each referred object and evaluating each prediction against its corresponding ground-truth annotation. Accordingly, we evaluate the localization of each referred object in such multi-object grounding setting.

We assess grounding performance in two setups: point estimation and bounding box-based estimation. As ALBEF, BLIP, and TinyLLaVA are not instruction-tuned to directly generate bounding boxes, we evaluate them using point estimation only, whereas Qwen2.5VL is evaluated using both point- and bounding-box estimation. For point estimation, we adopt Pointing Game Accuracy~\cite{pointgame}, which measures the percentage of predicted points that fall within the ground-truth bounding box. For bounding-box estimation, we report Acc@0.5.

For contrastive VLMs such as ALBEF and BLIP, we directly use the object query phrases as input. However, for the generative model TinyLLaVA, we construct input-output sequences for each query phrase as: \texttt{``Are there \{query phrase\} in the image? There are \{query phrase\} in the image.''} to obtain output-to-visual localization map. In all cases, localization maps for all queried objects are obtained in single forward pass. Object-specific point predictions are then computed by taking the peak attention from the localization map corresponding to each respective object phrase as detailed in Sec.~\ref{subsec:localization}. For Qwen2.5VL, we instruct the model to directly output either object point coordinates or bounding-box coordinates for all objects in the given query.

\vspace{0.1in}
\noindent\textbf{VQA.} 
Since generative-based VLMs are also instruction-tuned for VQA tasks, we 
evaluate their performance on two popular general VQA benchmarks SeedBench~\cite{seedbench} and SeedBench2~\cite{seedbench2} using their default evaluation protocols. 

\begin{table*}[t]
\centering
\scriptsize
\setlength{\tabcolsep}{3pt}
\renewcommand{\arraystretch}{1.15}

\caption{Single- and two-object grounding results based on Point Game Accuracy and Bounding-Box Acc@0.5. The numbers in \textbf{bold} indicate the best results in each group.
}
\label{tab:maingrounding}
\resizebox{\textwidth}{!}{
\begin{tabular}{l c>{\columncolor{myblue}}c c>{\columncolor{gray!20}}c c>{\columncolor{gray!20}}c c>{\columncolor{gray!20}}c c>{\columncolor{gray!20}}c}
\toprule
 & \multicolumn{2}{c}{\textbf{Average}}
 & \multicolumn{2}{c}{\bf Flickr30K}
 & \multicolumn{2}{c}{\bf RefCOCO}
 & \multicolumn{2}{c}{\bf RefCOCOg}
 & \multicolumn{2}{c}{\bf RefCOCO+} \\
\cmidrule(lr){2-3} \cmidrule(lr){4-5} \cmidrule(lr){6-7} \cmidrule(lr){8-9} \cmidrule(lr){10-11}

\textbf{No. of objects}
 & 1 & 2
 & 1 & 2
 & 1 & 2
 & 1 & 2
 & 1 & 2 \\
\midrule

TAS \cite{TAS}
& - & \cellcolor{white}-
& 77.9 & \cellcolor{white}-
& - & \cellcolor{white}-
& - & \cellcolor{white}-
& - & \cellcolor{white}- \\

G++ \cite{g++}
& - & \cellcolor{white}-
& 78.1 & \cellcolor{white}-
& - & \cellcolor{white}-
& - & \cellcolor{white}-
& - & \cellcolor{white}- \\

WWbL \cite{wwbl}
& - & \cellcolor{white}-
& - & \cellcolor{white}-
& 31.1 & \cellcolor{white}-
& 31.8 & \cellcolor{white}-
& 34.6 & \cellcolor{white}- \\

TRIS \cite{TRIS}
& - & \cellcolor{white}-
& - & \cellcolor{white}-
& 51.9 & \cellcolor{white}-
& 52.6 & \cellcolor{white}-
& 40.9 & \cellcolor{white}- \\

PCNet \cite{progresssive_comprehension}
& - & \cellcolor{white}-
& - & \cellcolor{white}-
& 60.6 & \cellcolor{white}-
& 57.7 & \cellcolor{white}-
& 56.5 & \cellcolor{white}- \\

\multicolumn{11}{l}{\em \tiny The above numbers are for point acc. and are for reference only; direct comparison may not be appropriate.}\\

\midrule

\multicolumn{11}{l}{Training data: MSCOCO}\\
\midrule

ALBEF \cite{ALBEF}
& 66.6 & 55.4
& 79.4 & 78.8
& 56.9 & 41.5
& \textbf{68.1} & 52.7
& 62.0 & 48.8 \\

\hspace{1em}+ SelfEQ~\cite{SelfEQ}
& 68.0 & 58.1
& 84.1 & 83.8
& 57.6 & 42.9 
& 67.6 & 54.8
& 62.5 & 51.1 \\
\hspace{1em}+ CompART (Ours)
& \textbf{68.1}\textcolor{mygreen}{$_\mathbf{\uparrow 1.5}$} 
& \textbf{61.7}\textcolor{mygreen}{$_\mathbf{\uparrow 6.3}$}
& \textbf{85.5} & \textbf{86.7}
& \textbf{58.2} & \textbf{49.1}
& 66.1 & \textbf{58.0}
& \textbf{62.7} & \textbf{53.2} \\

BLIP \cite{BLIP}
& 66.8 & 62.5
& 81.3 & 84.9
& 58.1 & 51.6
& 65.9 & 59.6
& 61.8 & 53.7 \\

\hspace{1em}+ CompART (Ours)
& \textbf{68.8}\textcolor{mygreen}{$_\mathbf{\uparrow 2.1}$} 
& \textbf{63.7}\textcolor{mygreen}{$_\mathbf{\uparrow 1.3}$}
& \textbf{84.7} & \textbf{86.8}
& \textbf{59.8} & \textbf{52.2}
& \textbf{67.5} & \textbf{61.1}
& \textbf{63.4} & \textbf{54.8} \\

TinyLLaVA-0.5B \cite{zhou2024tinyllava}
& 19.8 & 24.4
& 33.4 & 42.8
& 12.7 & 17.0
& 19.5 & 20.3
& 13.6 & 17.6 \\

\hspace{1em}+ CompART (Ours)
& \textbf{28.8}\textcolor{mygreen}{$_\mathbf{\uparrow 9.0}$} 
& \textbf{27.1}\textcolor{mygreen}{$_\mathbf{\uparrow 2.7}$}
& \textbf{44.3} & \textbf{44.2}
& \textbf{21.3} & \textbf{20.1}
& \textbf{26.1} & \textbf{22.8}
& \textbf{23.4} & \textbf{21.2} \\

\midrule
\multicolumn{11}{l}{Training data: Pixmo}\\
\midrule

ALBEF \cite{ALBEF}
& 67.1 & 55.9
& 80.3 & 79.3
& 57.3 & 42.0
& \textbf{68.4} & 53.1
& 62.5 & 49.1 \\

\hspace{1em}+ CompART (Ours)
& 67.1
& \textbf{57.0}\textcolor{mygreen}{$_\mathbf{\uparrow 1.2}$}
& \textbf{81.4} & \textbf{80.8}
& \textbf{57.7} & \textbf{42.4}
& 66.6 & \textbf{55.0}
& \textbf{62.7} & \textbf{49.9} \\

BLIP \cite{BLIP}
& 65.4 & 61.6
& 81.1 & \textbf{86.3}
& 56.2 & 50.1
& 65.0 & 57.9
& 59.2 & 52.1 \\

\hspace{1em}+ CompART (Ours)
& \textbf{66.8}\textcolor{mygreen}{$_\mathbf{\uparrow 1.5}$} 
& \textbf{62.5}\textcolor{mygreen}{$_\mathbf{\uparrow 0.8}$}
& \textbf{81.8} & 84.3
& \textbf{58.4} & \textbf{51.7}
& \textbf{65.4} & \textbf{59.7}
& \textbf{61.7} & \textbf{54.1} \\

TinyLLaVA-0.5B \cite{zhou2024tinyllava}
& 20.8 & 25.1
& 34.7 & 43.1
& 13.5 & 17.7
& 20.7 & 21.2
& 14.6 & 18.4 \\

\hspace{1em}+ CompART (Ours)
& \textbf{28.4}\textcolor{mygreen}{$_\mathbf{\uparrow 7.5}$} 
& \textbf{30.0}\textcolor{mygreen}{$_\mathbf{\uparrow 4.9}$}
& \textbf{45.3} & \textbf{48.3}
& \textbf{19.3} & \textbf{21.7}
& \textbf{28.0} & \textbf{26.5}
& \textbf{20.9} & \textbf{23.6} \\

Qwen2.5VL-3B \cite{bai2025qwen2}
& 89.3 & 72.3
& \textbf{88.6} & 82.9
& \textbf{92.9} & 77.1
& 90.2 & 59.9
& \textbf{85.5} & 69.2 \\

\hspace{1em}+ CompART (Ours)
& 89.2{$_{\downarrow 0.1}$}
& \textbf{74.9}\textcolor{mygreen}{$_\mathbf{\uparrow 2.6}$}
& 88.5 & \textbf{83.6}
& 92.8 & \textbf{80.6}
& 90.2 & \textbf{64.4}
& 85.3 & \textbf{71.1} \\


\midrule

Qwen2.5VL-3B$^{\text{BBox}}$ \cite{bai2025qwen2}
& 64.5 & 58.7
& 73.4 & 73.8
& 87.4 & 76.9
& 83.3 & 74.6
& \textbf{78.4} & 68.4 \\

\hspace{1em}+ CompART (Ours)
& 64.5
& \textbf{59.1}\textcolor{mygreen}{$_\mathbf{\uparrow 0.4}$}
& \textbf{73.5} & \textbf{74.2}
& 87.4 & \textbf{77.5}
& 83.3 & \textbf{75.2}
& 78.2 & \textbf{68.7} \\

\midrule

\multicolumn{11}{l}{\em \tiny These numbers are for bbox acc@0.5 and are for reference only; direct comparison may not be appropriate.}\\

WeakMCN \cite{cheng2025weakmcn}
& - & \cellcolor{white}-
& - & \cellcolor{white}-
& 67.2 & \cellcolor{white}-
& - & \cellcolor{white}-
& 44.4 & \cellcolor{white}- \\

ReCLIP \cite{reclip}
& - & \cellcolor{white}-
& - & \cellcolor{white}-
& 46.3 & \cellcolor{white}-
& 59.2 & \cellcolor{white}-
& 47.7 & \cellcolor{white}- \\

DeepSeek-VL-1.3B \cite{kang2025your}
& - & \cellcolor{white}-
& - & \cellcolor{white}-
& 73.9 & \cellcolor{white}-
& 67.3 & \cellcolor{white}-
& 61.9 & \cellcolor{white}- \\


DeepSeek-VL-7B \cite{kang2025your}
& - & \cellcolor{white}-
& - & \cellcolor{white}-
& 84.5 & \cellcolor{white}-
& 82.0 & \cellcolor{white}-
& 78.4 & \cellcolor{white}- \\

Shikra-7B \cite{chen2023shikra}
& - & \cellcolor{white}-
& - & \cellcolor{white}-
& 85.9 & \cellcolor{white}-
& 82.3 & \cellcolor{white}-
& 80.4 & \cellcolor{white}- \\

Ferret-7B \cite{you2023ferret}
& - & \cellcolor{white}-
& - & \cellcolor{white}-
& 87.1 & \cellcolor{white}-
& 84.4 & \cellcolor{white}-
& 80.4 & \cellcolor{white}- \\

\bottomrule
\end{tabular}
}
\vspace{-0.15in}
\end{table*}









\subsection{Results}
\label{subsec:multi-objectgrounding}

\noindent\textbf{Single- and Two-Object Grounding.}
In Table \ref{tab:maingrounding}, we present the results of single- and two-object grounding performance based on point game accuracy. To strictly evaluate the impact of our proposed CompART in improving the object grounding performance, we train the base VLMs using their original training objectives and compare them against models trained with CompART. 

As shown in Table \ref{tab:maingrounding}, CompART consistently improves grounding performance across all four benchmarks and model families. For example, ALBEF improves single-object and two-object grounding performance by 1.5\% and 6.7\% points, repectively, when trained with MSCOCO. Additionally, in single-object setting on TinyLLaVA-0.5B, CompART improves average performance by 9.0\% and 7.5\% points when trained with MSCOCO and PixMo, respectively. In the two-object setting, the corresponding improvements are 2.7\% and 4.9\% points.
Since, Qwen2.5VL-3B already achieves close to 90\% accuracy in the single-object setting, leaving limited room for improvement. In two-object setting, CompART improves grounding performance of  Qwen2.5VL-3B by an average of 2.6\% points. Since,  Qwen2.5VL-3B can directly produce object bounding boxes, we also evaluate their object grounding performance based on bounding box estimation. As shown in Table \ref{tab:maingrounding}, despite not explicitly regularizing object boundaries, CompART consistently improves two-object bounding-box grounding performance across all four benchmarks. 


Note that most existing grounding methods do not address multi-object grounding. Therefore, for reference, we report their single-object grounding results using both Pointing Game Accuracy and bounding-box Acc@0.5.
We also include a recent method, SelfEQ~\cite{SelfEQ}, that targets implicit grounding in contrastive VLMs through a similar lightweight training as ours. For a fair comparison, we recompute its performance across all benchmarks for both single- and two-object grounding. As shown in Table~\ref{tab:maingrounding}, our Qwen2.5VL-3B trained with CompART achieves comparable or superior performance, even when compared to larger VLMs such as DeepSeek-VL-7B, Shikra-7B, and Ferret-7B.

\begin{table}[t]
\centering
\scriptsize
\setlength{\tabcolsep}{3pt}
\renewcommand{\arraystretch}{1.15}
\caption{Evaluating generalization capability of CompART in three-, four-, and five-object grounding setups based on Point Game Accuracy and Bounding-Box Acc@0.5. The numbers in \textbf{bold} indicate the best results.
}
\label{tab:ablate_moreobjs}
\begin{tabular}{l ccc ccc ccc ccc}
\toprule
 & \multicolumn{3}{c}{\bf Flickr30K} 
 & \multicolumn{3}{c}{\bf RefCOCO}
 & \multicolumn{3}{c}{\bf RefCOCOg}
 & \multicolumn{3}{c}{\bf RefCOCO+} \\
 \cmidrule(lr){2-4} \cmidrule(lr){5-7} 
 \cmidrule(lr){8-10} \cmidrule(lr){11-13} 

\textbf{No. of objects}
 & 3 & 4 & 5 
 & 3 & 4 & 5 
 & 3 & 4 & 5 
 & 3 & 4 & 5 \\
\midrule

TinyLLaVA-0.5B
& 41.3 & 39.1 & 36.5 
& 13.7 & 11.5 & 10.1 
& 19.7 & 18.0 & 16.9 
& 15.0 & 12.4 & 11.7 \\

\hspace{1em}+CompART 
& \textbf{45.3} & \textbf{41.9} & \textbf{37.9} 
& \textbf{16.1} & \textbf{13.1} & \textbf{11.6} 
& \textbf{22.2} & \textbf{19.1} & \textbf{17.5} 
& \textbf{18.4} & \textbf{14.8} & \textbf{13.5} \\

\hspace{2em}$\Delta$
& \textcolor{mygreen}{+4.0} & \textcolor{mygreen}{+2.8} & \textcolor{mygreen}{+1.4} 
& \textcolor{mygreen}{+2.3} & \textcolor{mygreen}{+1.6} & \textcolor{mygreen}{+1.5} 
& \textcolor{mygreen}{+2.5} & \textcolor{mygreen}{+1.1} & \textcolor{mygreen}{+0.6} 
& \textcolor{mygreen}{+3.4} & \textcolor{mygreen}{+2.4} & \textcolor{mygreen}{+1.7} \\

\midrule
Qwen2.5VL-3B 
& 72.5 & 63.9 & 63.4 
& 67.1 & 64.0 & 59.8 
& 58.3 & 64.2 & 61.7 
& 57.9 & 52.1 & 43.1 \\

& \textbf{77.5} & \textbf{72.4} & \textbf{68.9} 
& \textbf{69.5} & \textbf{64.4} & \textbf{60.0} 
& \textbf{60.2} & \textbf{65.2} & \textbf{62.3} 
& \textbf{59.9} & \textbf{52.5} & \textbf{43.3} \\

\hspace{1em}+CompART 
& \textcolor{mygreen}{+5.0} & \textcolor{mygreen}{+8.5} & \textcolor{mygreen}{+5.5} 
& \textcolor{mygreen}{+2.4} & \textcolor{mygreen}{+0.4} & \textcolor{mygreen}{+0.3} &  \textcolor{mygreen}{+1.9} & \textcolor{mygreen}{+1.0} & \textcolor{mygreen}{+0.6} 
& \textcolor{mygreen}{+2.0} & \textcolor{mygreen}{+0.5} & \textcolor{mygreen}{+0.2} \\

\midrule
Qwen2.5VL-3B$^{\text{BBox}}$  
& 68.0 & 61.8 & 56.3 
& 66.2 & 57.0 & 51.3 
& 65.6 & 55.8 & 49.0 
& 56.5 & 44.9 & 37.1 \\

\hspace{1em}+CompART 
& \textbf{68.3} & \textbf{62.3} & \textbf{57.0} 
& \textbf{67.4} & \textbf{58.6} & \textbf{53.8} 
& \textbf{66.4} & \textbf{57.7} & \textbf{50.6} 
& \textbf{57.0} & 44.9 & \textbf{37.5} \\

\hspace{2em}$\Delta$
& \textcolor{mygreen}{+0.3} & \textcolor{mygreen}{+0.5} & \textcolor{mygreen}{+0.7} 
& \textcolor{mygreen}{+1.1} & \textcolor{mygreen}{+1.6} & \textcolor{mygreen}{+2.4} 
& \textcolor{mygreen}{+0.9} & \textcolor{mygreen}{+1.9} & \textcolor{mygreen}{+1.6} 
& \textcolor{mygreen}{+0.6} & 0.0 & \textcolor{mygreen}{+0.4} \\

\bottomrule
\end{tabular}
\vspace{-0.15in}
\end{table}

\noindent\textbf{Generalization beyond Two-Objects.} 
While CompART is trained to regularize attention over two objects, we stress-test its capabilities in grounding upto five-objects. The results presented in Table \ref{tab:ablate_moreobjs} show that CompART effectively generalizes well in all three-, four-, and five-object grounding settings across all four benchmarks. 
Specifically, for TinyLLaVA-0.5B, CompART yields large gains on Flickr30K, with 4.0\%, 2.8\%, and 1.4\% points for three-, four-, and five-object grounding, respectively.
For Qwen2.5VL-3B point estimation, we also observe significant gains on Flickr30K, with 5.0\%, 8.5\%, and 5.5\% points for three-, four-, and five-object grounding, respectively.
In the case of Qwen2.5VL-3B bounding box-based estimation, we observe increasing gains as the number of objects increases. For example, improvements of 1.1\%, 1.6\%, and 2.4\% points are observed for three-, four-, and five-object grounding on RefCOCO. 

\begin{table*}[t]
\centering
\scriptsize
\setlength{\tabcolsep}{3pt}
\caption{Results on VQA tasks using SeedBench and SeedBench2. We report accuracy, and the numbers in \textbf{bold} indicate the best results. 
}
\label{tab:seedbench}
\vspace{-0.15in}
\centering
\begin{tabular}{l cccc ccc ccc}
\toprule
 
 &
 \multicolumn{10}{c}{\textbf{SeedBench}}
 \\
\cmidrule(lr){2-11} 

& \textbf{Avg.} & Ident. & Attr. & Loc.
& Count & Spatial & Interact
& Visual & Text & Scene \\
\midrule 
Qwen2.5VL-3B
& 62.8 & 64.9 & 69.5 & 52.3
& 67.4 & 44.1 & \textbf{59.8}
& 62.8 & 78.3 & 66.2 \\

+CompART
& \textbf{65.0}\textcolor{mygreen}{$_\mathbf{+2.2}$}
& \textbf{66.8}
& \textbf{74.6}
& \textbf{56.9}
& \textbf{68.7}
& \textbf{47.2}
& 58.8
& \textbf{64.7}
& 78.3
& \textbf{69.2} \\

\midrule



 &
 \multicolumn{10}{c}{\textbf{SeedBench2}}
 \\ \cmidrule(lr){2-9}

& \textbf{Avg.}
& Celeb. & Land. & Chart.
& RefExp. & Sci.
& Emotion & Math.  \\
\cmidrule(lr){2-9}

Qwen2.5VL-3B
& 58.5
& 74.2 & 75.2 & 81.8
& 43.7 & 74.4
& 13.0 & 47.0 \\

+CompART
& \textbf{60.5}\textcolor{mygreen}{$_\mathbf{\uparrow 2.0}$}
& \textbf{75.8}
& \textbf{76.8}
& \textbf{82.4}
& \textbf{46.2}
& 74.4
& \textbf{18.8}
& \textbf{49.2} \\

\bottomrule
\end{tabular}

\end{table*}

\vspace{0.1in}
\noindent\textbf{General VQA.}
To evaluate the impact of improved multi-object grounding on VQA performance, we evaluate Qwen2.5VL-3B on SeedBench~\cite{seedbench} and SeedBench2~\cite{seedbench2}. These benchmarks cover a diverse set of vision tasks, including single- and multi-object understanding, spatial relation, and chart understanding, among others. 
As shown in Tables~\ref{tab:seedbench}, CompART improves overall performance by $2.2\%$ and $2.0\%$ on SeedBench and SeedBench2, respectively. Notably, CompART yields large gains on single-object tasks, improving performance by 5.1\% and 4.6\% points on instance attribute and instance location tasks. We also observe substantial  improvements of 3.1\% and 3.0\% points on spatial relation and scene understanding tasks. Finally, on SeedBench2, CompART improves performance by 2.5\% points on both expression and emotion recognition tasks, and by 2.2\% points on math-related VQA tasks.

\subsection{Ablation Study \& Analysis}
\label{subsec:ablation}
\noindent


\begin{table}[t]
\centering
\scriptsize
\caption{\textbf{Ablation on Loss.} We present grounding performance for one- to three-object settings using Pointing Game Accuracy. 
$^\dagger$ refers to the VLM variant trained with original image captions instead of phrase chunks. 
Pixmo is used for training. The numbers in \textbf{bold} indicate the best results in each group.
}
\vspace{-0.15in}
\setlength{\tabcolsep}{4pt}
\begin{tabular}{@{}l c ccc ccc ccc@{}}
\toprule
& \multicolumn{3}{c}{\bf RefCOCO} & \multicolumn{3}{c}{\bf RefCOCOg} & \multicolumn{3}{c}{\bf RefCOCO+} \\
\cmidrule(lr){2-4}\cmidrule(lr){5-7}\cmidrule(lr){8-10} 
\textbf{No. of objects}
& 1 & 2 & 3
& 1 & 2 & 3
& 1 & 2 & 3 \\
\midrule

TinyLLaVA-0.5B 
& 12.7 & 17.0 & 13.3
& 19.5 & 20.3 & 18.9
& 13.6 & 17.6 & 14.1 \\

\hspace{0.8em} + $\mathcal{L}_{\text{base}}^\dagger$
& 13.5 & 17.7 & 13.7
& 20.7 & 21.2 & 19.7
& 14.6 & 18.4 & 15.0 \\

\hspace{0.8em} + $\mathcal{L}_{\text{base}}$
& 13.7 & 16.9 & 13.7
& 20.4 & 22.4 & 21.0
& 15.3 & 18.6 & 15.3 \\

\hspace{0.8em} + $\mathcal{L}_{\text{base}}$ + $\mathcal{L}_{\text{Comp}}$ 
& \textbf{19.3} & \textbf{21.7} & \textbf{16.1}
& \textbf{28.0} & \textbf{26.5} & \textbf{22.2}
& \textbf{20.9} & \textbf{23.6} & \textbf{21.7} \\






\bottomrule
\end{tabular}
\label{tab:ablation}
\vspace{-0.15in}
\end{table}

\noindent \textbf{Ablation on Loss}: 
We conduct a thorough ablation exploring the impact of individual components of CompART. The results presented in Table~\ref{tab:ablation} show that simply training the base VLM with the image-caption pairs ($\mathcal{L_\text{base}}^\dagger$) or image-phrase chunks ($\mathcal{L_\text{base}}$) does not lead to any noticeable improvements over the base model. However, training with the $\mathcal{L_\text{Comp}}$ loss significantly improves grounding performance, with average gains of $6.3\%$, $4.6\%$, and $3.3\%$ in the one-, two-, and three-object settings, respectively.



\begin{figure}
    \centering
    \begin{tabular}{cc}
         \includegraphics[width=0.48\linewidth]{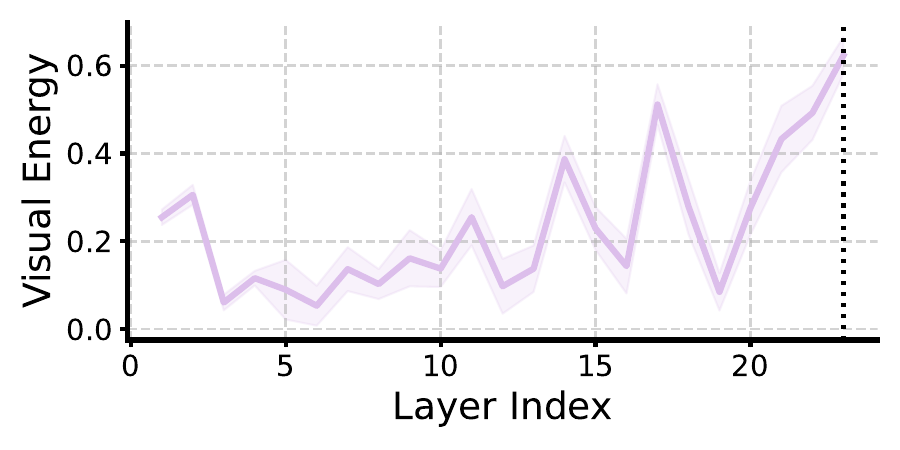}
         & 
         \includegraphics[width=0.48\linewidth]{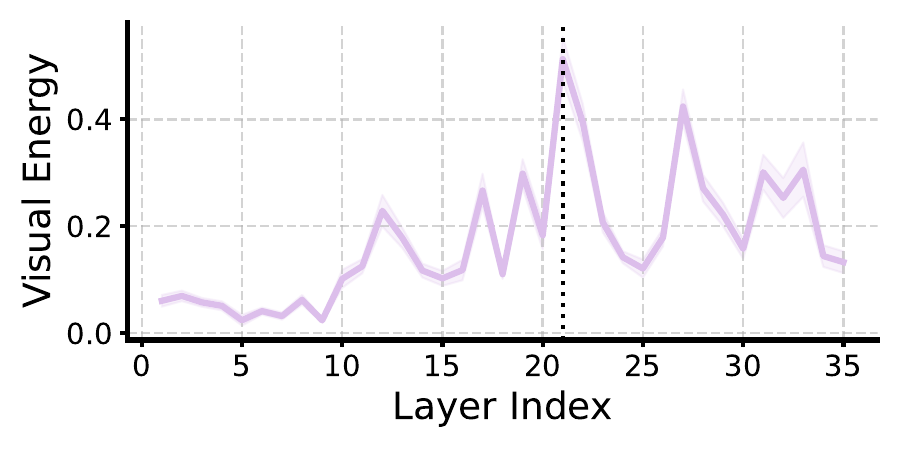}
    \end{tabular}

    \vspace{-0.15in}
    \caption{
    We present the total output-to-visual attention across all layers for TinyLLaVA-0.5B (left) and Qwen2.5VL-3B (right). The layers exhibiting the highest visual attention are indicated by vertical dashed lines; i.e., the $23^{\text{rd}}$ layer in TinyLLaVA-0.5B and the $21^{\text{st}}$ layer in Qwen2.5VL-3B.
    }
    \label{fig:attn_layer}
    \vspace{-0.2in}
\end{figure}


\noindent
\textbf{Layer Section.}
To compute object-specific phrase-localization maps, we use the output-to-visual attention from the VLM layers exhibiting the highest visual attention. To estimate, we randomly sample 500 images from MSCOCO and compute the output-to-visual attention across all the layers for both TinyLLaVA-0.5B and Qwen2.5VL-3B. As shown in Figure~\ref{fig:attn_layer}, the $23^{\text{rd}}$ layer in TinyLLaVA-0.5B exhibits the highest attention, while the $21^{\text{st}}$ layers in Qwen2.5VL-3B show the highest attention energy. Accordingly, we use the attention from the respective layers for TinyLLaVA-0.5B and Qwen2.5VL-3B to compute the corresponding phrase-localization maps.

\begin{figure}[!bp]
\vspace{-0.15in}
    \centering
    \includegraphics[width=1\linewidth]{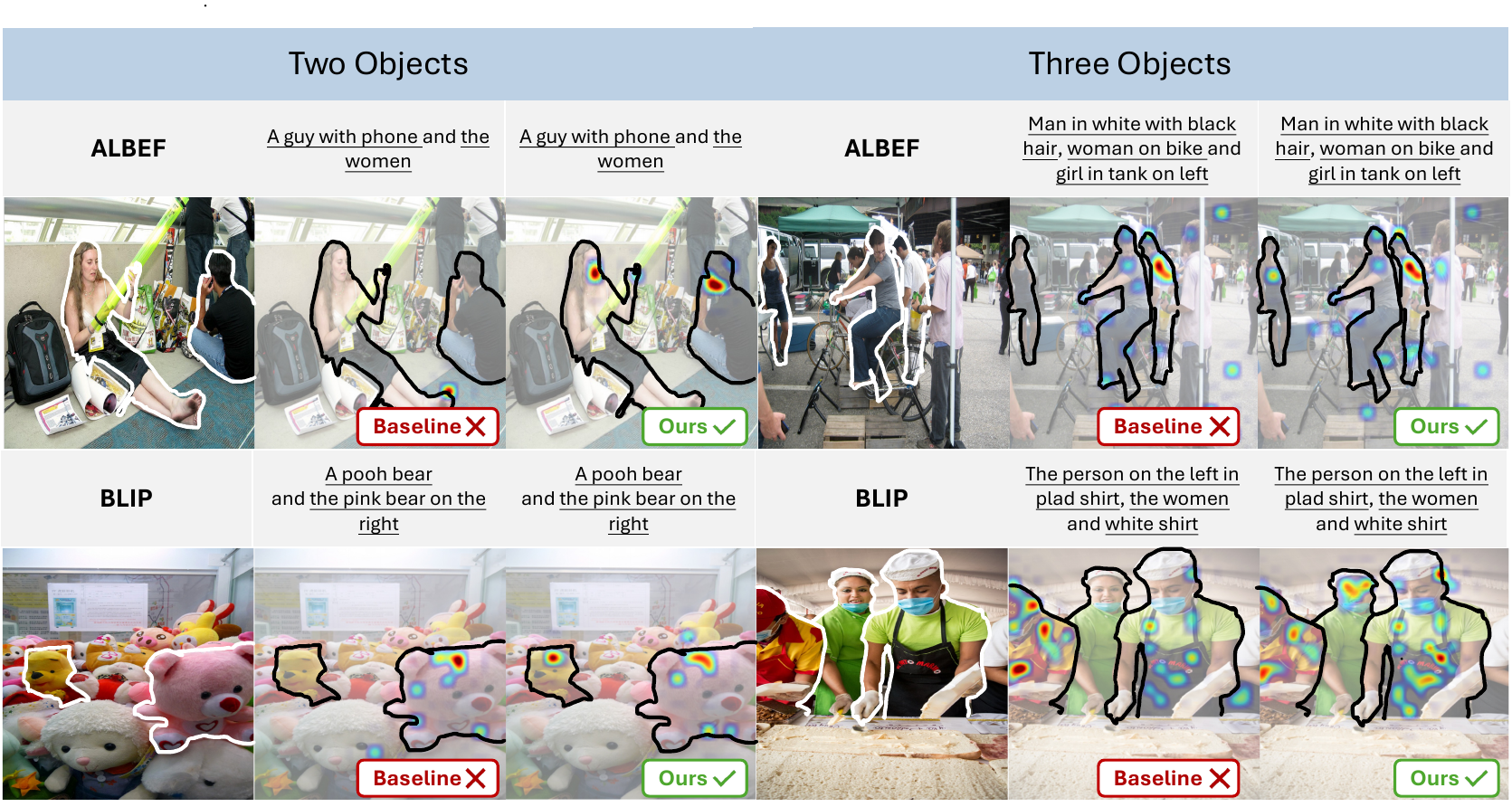}
    \caption{\textbf{Qualitative results for contrastive models.} White/black outlines indicate the ground-truth referred objects. With CompART, the models produce improved attention maps that ground multiple referred objects simultaneously.
    }
    \label{fig:quali1}
    \vspace{-15pt}
\end{figure}

\begin{figure}[!t]
    \centering
    \includegraphics[width=1\linewidth]{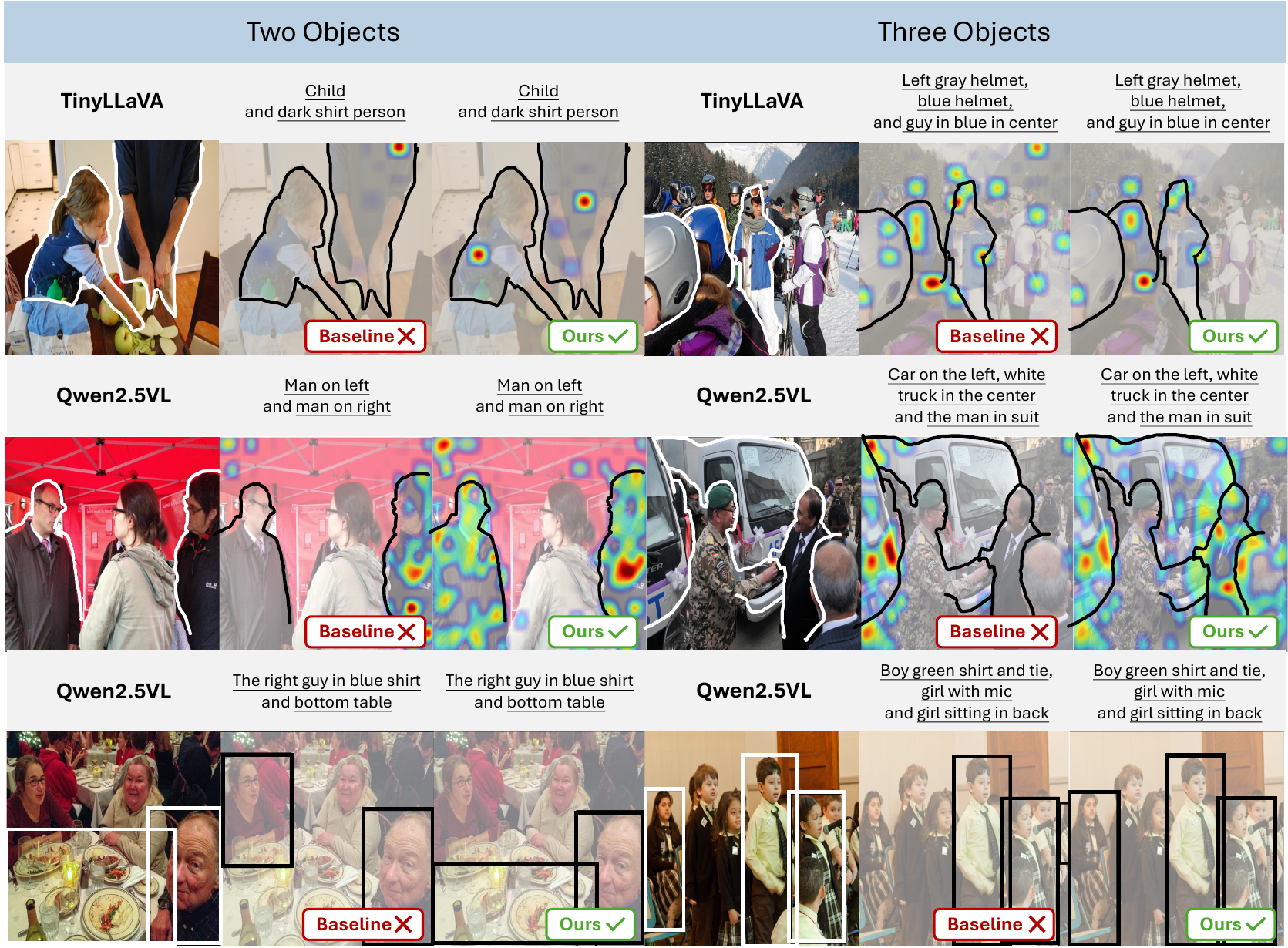}
    \caption{\textbf{Qualitative results for generative models.} White/black outlines indicate the ground-truth referred objects. With CompART, the models produce improved attention maps that ground multiple referred objects simultaneously. Unlike TinyLLaVA, Qwen2.5VL can directly produce bounding boxes for the queried objects; therefore, we present qualitative analysis for both point estimation and bounding box estimation.
    }
    \label{fig:quali2}
    \vspace{-0.15in}
\end{figure}

\vspace{-10pt}
\subsection{Qualitative Results}

Figure \ref{fig:quali1} and Figure \ref{fig:quali2} present qualitative analyses for both contrastive and generative VLMs. The results show that base VLMs fail to correctly localize all regions corresponding to multiple objects in a query, and often attention collapses to a single region. In contrast, CompART substantially improves object grounding performance with improved attention, 
not only in the two-object setting—on which the model is trained—but also generalizes beyond this setting to more complex scenarios involving three-object grounding.

\section{Conclusion}
\label{sec:conclusion}

In this paper, we identify a consistent performance gap where both contrastive and generative VLMs, despite succeeding at single-object localization, struggle to attend to multiple entities simultaneously when presented with composite queries. This attention collapse limits the utility of VLMs in complex, real-world vision tasks. To mitigate this, we propose CompART, a lightweight and architecture-agnostic method that leverages the inherent compositionality of natural language as supervision. By decomposing captions into object-centric phrases and regularizing the model's attention maps through our proposed \emph{composition loss}, we encourage models to treat composite references as the sum of their constitutes. Our results across four diverse architectures and four benchmarks demonstrate that CompART significantly improves grounding accuracy for multi-object queries without requiring additional dense annotations. Furthermore, the improved grounding leads to better performance on diverse downstream VQA tasks. We hope this work encourages further exploration into improving multi-object visual grounding.


\clearpage  


%
%
\bibliographystyle{splncs04}
\bibliography{main}

\clearpage
\appendix
\renewcommand{\thesection}{\Alph{section}}
\renewcommand{\theHsection}{appendix.\Alph{section}}

\startcontents[app]

\setcounter{table}{0}
\setcounter{figure}{0}
\setcounter{equation}{0}
\setcounter{page}{1}

\renewcommand{\thetable}{\Alph{section}\arabic{table}}
\renewcommand{\thefigure}{\Alph{section}\arabic{figure}}
\renewcommand{\theequation}{\Alph{section}\arabic{equation}}

\begin{center}
\Large
Supplementary Material
\end{center}


\noindent
The organization of the supplementary material is as follows:
\begin{itemize}[noitemsep,nolistsep,leftmargin=48pt,label={Appendix }]
    \item \ref{supsec:eval} - Evaluation Details 
    \item \ref{supsec:training} - Training Details
    \item \ref{supsec:results} - Additional Results
    
\end{itemize}

\section{Evaluation Details} \label{supsec:eval}

\subsection{Benchmark Statistics}

For grounding, we evaluate CompART on four visual grounding benchmarks: Flickr30K~\cite{flickr30k}, along with all validation and test splits of RefCOCO~\cite{refcoco+}, RefCOCO+~\cite{refcoco+}, and RefCOCOg~\cite{refcocog}. 
We assess performance across single- to five-object grounding settings. For single-object grounding, we use the full evaluation splits. For multi-object grounding, we consider images that contain ground-truth annotations for multiple objects and evaluate all possible object combinations.  The number of samples in each of the evaluation sets for multi-object grounding is summarized in Table \ref{tab:multiobj_counts}.

\begin{table}[!]
\caption{Number of evaluation samples across datasets. 
}
\centering
\scriptsize
\setlength{\tabcolsep}{6pt}
\renewcommand{\arraystretch}{1.15}

\begin{tabular}{cccccccccc}
\toprule

\multirow{2}{*}{
\specialcell{\bf Number of\\\bf Objects}
}
 & \textbf{Flickr} 
 & \multicolumn{3}{c}{\textbf{RefCOCO}}
 & \multicolumn{2}{c}{\textbf{RefCOCOg}}
 & \multicolumn{3}{c}{\textbf{RefCOCO+}} \\
 \cmidrule(lr){2-2} 
 \cmidrule(lr){3-5} 
 \cmidrule(lr){6-7}
 \cmidrule(lr){8-10} 

 &  Test
 & Test A & Test B & Val 
 & Test & Val
 & Test A & Test B & Val \\

\midrule
2 & 3900 & 2056 & 1715 & 3906 & 3669 & 1967 & 2056 & 1669 & 3884 \\
3 & 1909 & 1328 & 1193 & 2775 & 1835 & 1063 & 1328 & 1065 & 2739 \\
4 & 1740 & 755  & 1052 & 1972 & 854  & 577  & 755  & 744  & 1937 \\
5 & 1469 & 366  & 1040 & 1229 & 371  & 318  & 366  & 485  & 1208 \\
\bottomrule
\end{tabular}

\label{tab:multiobj_counts}
\end{table}



\noindent For VQA, we evaluate CompART on two benchmarks: SeedBench and SeedBench2. Specifically, for SeedBench, we evaluate on nine tasks with 14K samples in total, including Scene Understanding, Instance Identity, Instance Attributes, Instance Location, Instance Counting, Spatial Relation, Instance Interaction, Visual Reasoning, and Text Understanding. In the case of SeedBench2, we evaluate on the following seven tasks with 2.4K samples in total, including Celebrity Recognition, Landmark Recognition, Chart Understanding, Visual Referring Expression, Science Knowledge, Emotion Recognition, and Visual Mathematics.

\subsection{Multi-object Grounding Evaluation Setup}

In order to capture the extent to which the models simultaneously localize queried objects in multi-object grounding, we predict distinct points or bounding boxes for each referred object and evaluate each prediction against its corresponding ground-truth annotation as detailed below.

\noindent\textbf{Contrastive-based Models:} 
Since ALBEF~\cite{ALBEF} and BLIP~\cite{BLIP} are not equipped with directly predicting object coordinates, we instead derive them from 
cross-attention maps.
Specifically, we construct a query 
\(\{o_1, o_2, \ldots o_n\}\) with \(n\) target object phrases.
Next, we extract the text-to-image cross-attention map $\mathbf{A} \in \mathbb{R}^{(K_1+\dots +K_n) \times P}$, where $K_i$ denotes the corresponding token length of target $o_i$. 
The localization map for $o_i$ is obtained by averaging over its tokens:
$\mathbf{M}_i =
\frac{1}{K_i}\sum_{k=1}^{K_i}\!\left(
\mathbf{A}^{(k)}
\right)$. 
Then, we project $\mathbf{M}_i \in \mathbb{R}^{P}$ as $\mathbf{M}_i \in \mathbb{R}^{H \times W}$ where $H$ and $W$ denote the height and width of the image.
Finally, we compute the point coordinates of $o_i$ as the location of the maximum value in the localization map: 
\begin{equation}
    (h_i, w_i) = \arg\max_{(h, w)} \mathbf{M}_i[h, w], \quad h \in [H],\, w \in [W], 
\label{equation:maxpoint}
\end{equation}
where $h$ and $w$ correspond to the height and width indices of the pixels in $v$.

\noindent\textbf{Generative-based Models: } 
For a given set of $n$ target object phrases $\{o_1, \ldots, o_n\}$, we construct the query 
\texttt{``Are there \{$o_1$, \ldots, $o_n$\} in the image?''} 
and set the target response as 
\texttt{``There are \{$o_1$, \ldots, $o_n$\} in the image.''} 
Next, we extract the output-to-visual attention maps 
$\mathbf{A} \in \mathbb{R}^{(K_1 + \dots + K_n) \times P}$, 
corresponding to the tokens of the object phrases. 
Following the procedure described above for contrastive-based models, we compute the object-specific localization map for each $o_i$ as $\mathbf{M}_i = \frac{1}{K_i} \sum_{k=1}^{K_i} \mathbf{A}^{(k)}$.
The predicted point coordinates for each $o_i$ are then obtained in the same manner as in Eq.~\ref{equation:maxpoint}. For Qwen2.5-VL~\cite{bai2025qwen2}, which is instruction-tuned for coordinate prediction, we directly prompt the model to output object point or bounding box coordinates.


\section{Training Details}\label{supsec:training}

We apply CompART on ALBEF~\cite{ALBEF}, BLIP~\cite{BLIP}, TinyLLaVA-0.5B, and Qwen2.5VL-3B and train all the model with Pixmo~\cite{deitke2025molmo} and ALBEF~\cite{ALBEF}, BLIP~\cite{BLIP}, TinyLLaVA-0.5B with MSCOCO~\cite{COCO}. 

\subsection{Training Setup}

\noindent\textbf{ALBEF.} ALBEF combines a ViT-B image encoder and a BERT$_{\text{base}}$ text encoder. We train ALBEF with an image resolution of $256 \times 256$, using a learning rate of $2\times10^{-5}$ for COCO and $2\times10^{-6}$ for Pixmo. For the composition loss, we use the head-averaged Grad-CAM map from the third cross-attention layer. For COCO, the model is trained with a curriculum of 
$\mathcal{L}_{\text{base}} + \lambda \alpha \mathcal{L}_{\text{comp}}$,
where $\alpha$ starts at 1, decrease to 0 by the second epoch, and remains constant afterwards.
For Pixmo, which ALBEF is not pretrained on, we first train the model on Pixmo phrase chunks for 10 epochs, and then retrain it with the same curriculum as used for COCO.

\noindent\textbf{BLIP.}
We train BLIP$_{\text{large}}$, which adopts ViT-L/16 as the image encoder and BERT as the text encoder, with an additional cross-attention layer inserted in each transformer block of BERT. The model is trained with an image resolution of $224 \times 224$ and a learning rate of $1\times10^{-5}$ for both COCO and Pixmo. For the composition loss, we use the head-averaged Grad-CAM map from the ninth cross-attention layer. For COCO, we use the same training curriculum as in ALBEF
$\mathcal{L}_{\text{base}} + \lambda \alpha \mathcal{L}_{\text{comp}}$,
where $\alpha$ starts at 1, decrease to 0 by the second epoch, and remains constant afterwards.
For Pixmo, which BLIP is not pretrained on, we first train the model on Pixmo phrase chunks for 10 epochs and then retrain using the same curriculum mentioned above.

\noindent\textbf{TinyLLaVA.}
TinyLLaVA-0.5B uses SigLIP~\cite{siglip} as the visual encoder and Qwen2-0.5B~\cite{yang2024qwen2technicalreport} as the language model. The model is trained with an image resolution of $224 \times 224$, using a learning rate of $2\times10^{-5}$ for COCO and $2\times10^{-7}$ for Pixmo. For the composition loss, we use the head-averaged output-to-visual attention map from the 23$^\text{rd}$ self-attention layer. For COCO, on which the model has already been pretrained, we set $\alpha$ to 1 and train the model with $\mathcal{L}_{\text{base}} + \lambda \alpha \mathcal{L}_{\text{total}}$ for one epoch. For Pixmo, which the model has never been trained on, we train with $\mathcal{L}_{\text{base}} + \lambda \alpha \mathcal{L}_{\text{comp}}$ for a total of 2 epochs, where $\alpha$ is set to 0 for the first epoch and increase to 1 for the 2nd epoch.

\noindent\textbf{Qwen2.5-VL.}
Qwen2.5VL-3B employs a Qwen-specific Vision Transformer as the visual encoder and the Qwen2.5 LLM as the language model. We train Qwen2.5VL-3B using the native input resolution and a learning rate of $2\times10^{-7}$ for Pixmo. For the composition loss, we use the head-averaged output-to-visual attention map from the 21$^\text{st}$ self-attention layer. Similar to TinyLLaVA, we train using a curriculum of 
$\mathcal{L}_{\text{base}} + \lambda \alpha \mathcal{L}_{\text{comp}}$ for a total of 2 epochs,
where $\alpha$ is set to 0 for the first epoch and increase to 1 for the 2nd epoch.

\subsection{Phrase Chunking Setup}
\begin{figure}
    \centering
    \includegraphics[width=1\linewidth]{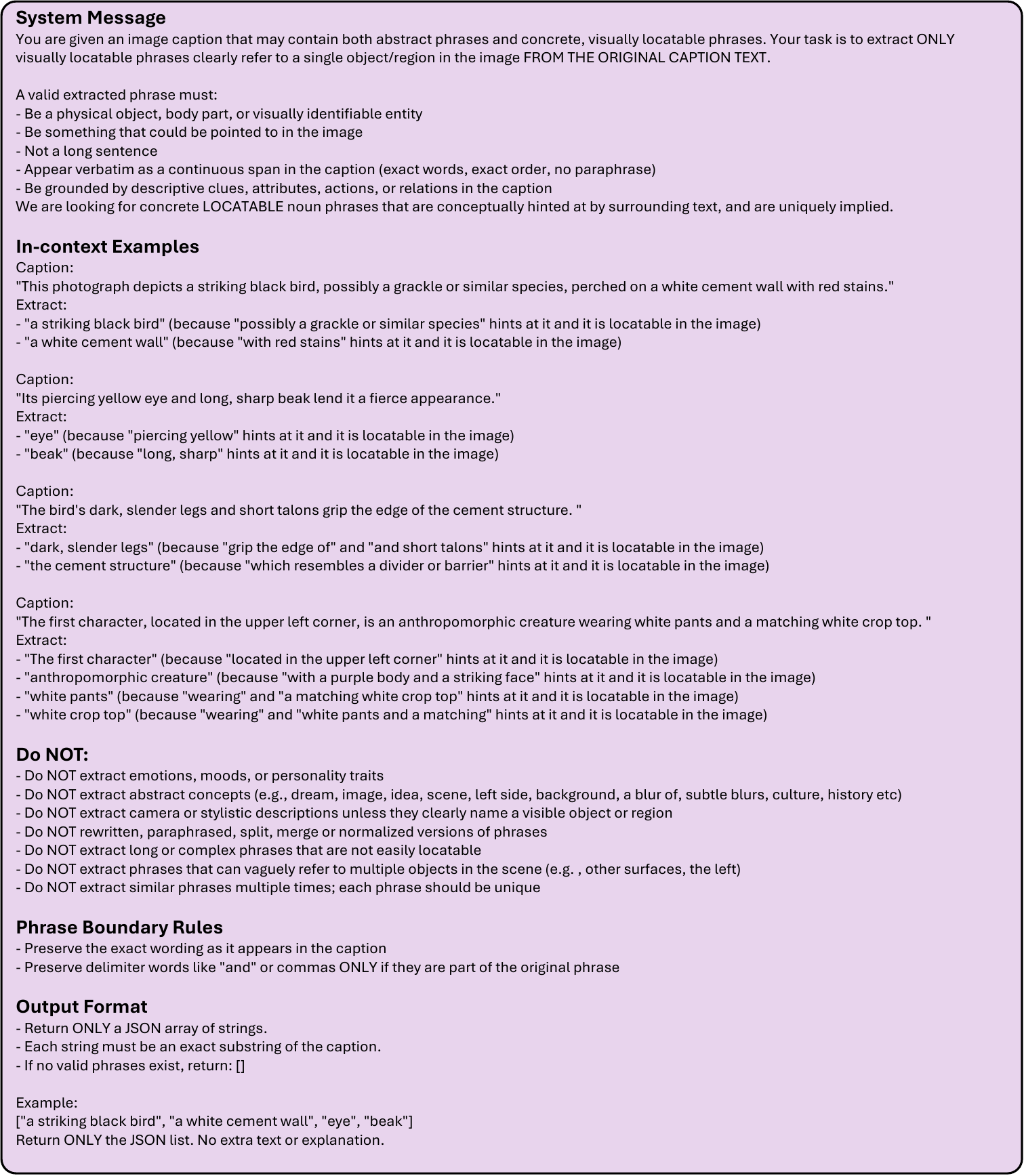}
    \caption{Instructions used with Qwen3-30B for phrase chunking on Pixmo captions}
    \label{fig:incontext}
\end{figure}

We use Qwen3-30B to do phrase chunking on Pixmo captions.  In Figure~\ref{fig:incontext}, we show the instruction, which includes in-context examples with explanations that guide the model to extract object-centric phrases.

We perform a human sanity check on Qwen3-30B phrase chunking for Pixmo captions. We randomly sample 180 phrase chunks and ask annotators whether each phrase refers to a concrete, visually locatable object or region in the image (Yes, No, or Unclear). Each phrase is annotated by three independent annotators. Across all annotations, 82.96\% of labels are ``Yes'', 7.78\% are ``No'', while the remaining annotations are labeled as unclear. At the phrase level, 155 out of 180 phrases (86.1\%) receive a majority “Yes” judgment, indicating that most generated chunks correspond to visually grounded entities. This proportion is significantly above chance (binomial test, $p < 10^{-23}$). Inter-annotator agreement is also substantial, with an average pairwise agreement of 0.89 and Krippendorff’ alpha \cite{krippendorff2018content} as  0.63. These results suggest that Qwen3-30B reliably produces phrase chunks that correspond to visually locatable entities in the image. Figure \ref{fig:humanstudy} shows the interface for collecting human annotations.

\begin{figure}
    \centering
    \includegraphics[width=1\linewidth]{figure/Humanstudy.pdf}
    \caption{Interface for collecting human evaluations of phrase chunk quality generated by Qwen3-30B}
    \label{fig:humanstudy}
\end{figure}

\noindent \section{Additional Results} \label{supsec:results}

\subsection{VQA}

We share additional VQA results of Gemma3-4B, InternVL3.5-4B and SmolVLM2.2B on SeedBench and SeedBench2. While our base VLM Qwen2.5-VL-3B performs similarly to InternVL3.5-4B and SmolVLM-2.2B, Qwen2.5-VL-3B + CompART outperforms these models. 

\begin{table*}[!]
\centering
\scriptsize
\setlength{\tabcolsep}{3pt}
\caption{Results on VQA tasks using SeedBench and SeedBench2. We report accuracy, and the numbers in \textbf{bold} indicate the best results. 
}
\label{supptab:seedbench}
\vspace{-0.15in}
\centering
\begin{tabular}{l cccc ccc ccc}
\toprule
 
 &
 \multicolumn{10}{c}{\textbf{SeedBench}}
 \\
\cmidrule(lr){2-11} 

& \textbf{Avg.} & Ident. & Attr. & Loc.
& Count & Spatial & Interact
& Visual & Text & Scene \\

Gemma3-4B~\cite{gemma3}
& 53.2 & 56.7 & 59.9 & 42.3
& 52.6 & 39.1 & 52.6
& 54.1 & 63.5 & 57.8 \\
LLaVA1.5-7B~\cite{llava1_5}
& 58.0 & 61.7 & 66.2 & 49.1
& 58.5 & 38.5 & 52.6
& 64.4 & 67.1 & 63.8 \\
InternVL3.5-4B~\cite{internvl35}
& 62.2 & 62.4 & 67.8 & 53.6
& 68.8 & 47.3 & 60.8
& 63.4 & 70.6 & 65.5 \\
SmolVLM-2.2B~\cite{smolvlm}
& 62.6 & 64.0 & 73.4 & 56.0
& 63.4 & 41.9 & 54.6
& 64.1 & 76.5 & 69.6 \\
\midrule 
Qwen2.5VL-3B
& 62.8 & 64.9 & 69.5 & 52.3
& 67.4 & 44.1 & \textbf{59.8}
& 62.8 & 78.3 & 66.2 \\

+CompART
& \textbf{65.0}\textcolor{mygreen}{$_\mathbf{+2.2}$}
& \textbf{66.8}
& \textbf{74.6}
& \textbf{56.9}
& \textbf{68.7}
& \textbf{47.2}
& 58.8
& \textbf{64.7}
& 78.3
& \textbf{69.2} \\

\midrule



 &
 \multicolumn{10}{c}{\textbf{SeedBench2}}
 \\ \cmidrule(lr){2-9}

& \textbf{Avg.}
& Celeb. & Land. & Chart.
& RefExp. & Sci.
& Emotion & Math.  \\
\cmidrule(lr){2-9}

Gemma3-4B~\cite{gemma3}
& 32.6
& 58.5 & 44.4 & 26.4
& 26.1 & 24.2
& 17.4 & 31.1 \\
LLaVA1.5-7B~\cite{llava1_5}
& 39.7
& 61.5 & 46.2 & 24.2
& 45.7 & 45.1
& 28.5 & 26.5 \\
InternVL3.5-4B~\cite{internvl35}
& 50.1
& 47.6 & 47.8 & 76.1
& 41.7 & 71.8
& 19.2 & 46.2 \\
SmolVLM-2.2B~\cite{smolvlm}
& 52.5
& 58.2 & 55.4 & 45.5
& 45.7 & 67.9
& 56.5 & 38.6 \\
\midrule

Qwen2.5VL-3B
& 58.5
& 74.2 & 75.2 & 81.8
& 43.7 & 74.4
& 13.0 & 47.0 \\

+CompART
& \textbf{60.5}\textcolor{mygreen}{$_\mathbf{\uparrow 2.0}$}
& \textbf{75.8}
& \textbf{76.8}
& \textbf{82.4}
& \textbf{46.2}
& 74.4
& \textbf{18.8}
& \textbf{49.2} \\

\bottomrule
\end{tabular}

\end{table*}

\end{document}